\documentclass[10pt,twocolumn,letterpaper]{article}

\usepackage{cvpr}
\usepackage{times}
\usepackage{epsfig}
\usepackage{graphicx}
\usepackage{amsmath}
\usepackage{amssymb}

\usepackage{comment}
\usepackage{xcolor}
\usepackage{multirow}
\usepackage{booktabs}
\usepackage{subcaption}
\usepackage{bbold}
\usepackage{textcomp}
\usepackage{pifont}
\usepackage{mathtools}

\definecolor{ColorGray}{RGB}{128, 128, 128}
\definecolor{ColorDarkGreen}{RGB}{19, 179, 50}
\definecolor{ColorLightBlue}{RGB}{18, 137, 255}

\newcommand{\Figure}[1]{Fig.~#1}
\newcommand{\Table}[1]{Table~#1}
\newcommand{\Equation}[1]{Eq.~#1}
\newcommand{\Section}[1]{Sec.~#1} 
\newcommand{\Etal}[1]{#1 et al.}

\def\AuthorMargin{\hspace{1cm}}
\def\InstitutionMargin{\hspace{0.8cm}}

\def\MVDescExtractor{f}

\def\PointP{\mathbf{p}}
\def\PointQ{\mathbf{q}}
\def\PointCloudP{\mathcal{P}}
\def\PointCloudQ{\mathcal{Q}}
\def\PointProperty{\mathcal{C}}

\def\ProbabilityMap{\mathcal{D}}
\def\AggregationWeight{w}
\def\PointDepth{z}

\def\ViewPoint{\mathbf{c}}
\def\ViewPointTheta{\theta}
\def\ViewPointPhi{\phi}
\def\ViewPointRadius{\rho}
\def\ViewPointUp{\mathbf{u}}
\def\ViewNum{n}
\def\ViewImage{\mathcal{I}}
\def\ViewFeatureMap{\mathcal{F}}
\def\FeatureFusion{\tilde{\mathcal{F}}}
\def\FeatureFusionWeight{\alpha}

\def\MVDescDim{d}

\def\PointPairSet{\mathcal{B}}
\def\Loss{\mathcal{L}}
\def\TripletMargin{m}

\def\FragmentPairSet{\mathcal{G}}
\def\MatchPointPairSet{\mathcal{M}}
\def\DescExtractor{g}
\def\MatchRecall{\mathcal{R}}
\def\MatchTransform{\mathcal{T}}

\usepackage[pagebackref=true,breaklinks=true,letterpaper=true,colorlinks,bookmarks=false]{hyperref}

\cvprfinalcopy 

\def\cvprPaperID{****} 

\begin{document}

\title{End-to-End Learning Local Multi-view Descriptors for 3D Point Clouds}

\author{Lei Li$^{1}$\thanks{L. Li was an intern at Alibaba A.I. Labs.}\AuthorMargin{}Siyu Zhu$^{2}$\AuthorMargin{}Hongbo Fu$^{3}$\thanks{H. Fu is the corresponding author. E-mail: hongbofu@cityu.edu.hk}\AuthorMargin{}Ping Tan$^{2,4}$\AuthorMargin{}Chiew-Lan Tai$^{1}$\\
	${}^{1}$HKUST\InstitutionMargin{}${}^{2}$Alibaba A.I. Labs\InstitutionMargin{}${}^{3}$City University of Hong Kong\InstitutionMargin{}${}^{4}$Simon Fraser University 
}

\maketitle

\begin{abstract}
	
	In this work, we propose an end-to-end framework to learn local multi-view descriptors for 3D point clouds. To adopt a similar multi-view representation, existing studies use hand-crafted viewpoints for rendering in a preprocessing stage, which is detached from the subsequent descriptor learning stage. In our framework, we integrate the multi-view rendering into neural networks by using a differentiable renderer, which allows the viewpoints to be optimizable parameters for capturing more informative local context of interest points. To obtain discriminative descriptors, we also design a soft-view pooling module to attentively fuse convolutional features across views. Extensive experiments on existing 3D registration benchmarks show that our method outperforms existing local descriptors both quantitatively and qualitatively.
	
\end{abstract}


\section{Introduction}
\label{sec:introduction}

Local descriptors for 3D geometry are widely recognized as one of the cornerstones in many computer vision and graphics tasks, such as correspondence establishment, registration, segmentation, retrieval, etc.
Particularly, with the prevalence of consumer-level RGB-D sensors, voluminous scanned data requires robust local descriptors for scene alignment and reconstruction~\cite{Xiao_2013_ICCV,Choi_2015_CVPR}.
Such 3D data, however, is often noisy and incomplete, presenting challenges to the design of local descriptors.

Existing hand-engineered local descriptors~\cite{Johnson:1999:SPIN,Frome:2004:3DShapeContext,Rusu:2008:PFH,Rusu:2009:FPFH,Tombari:2010:USH,Tombari:2010:USC:1877808.1877821,Salti:2014:SHOTUS}, proposed in the past few decades, are mostly built upon histograms of low-level 3D geometric properties.
Recent trends with deep neural networks have motivated researchers to develop learning-based local descriptors in a data-driven manner~\cite{Zeng_2017_CVPR,Elbaz:2017:PCRL,Khoury_2017_ICCV,Huang:2017:LLS:3151031.3137609,Deng_2018_CVPR,Wang_2018_ECCV,Gojcic_2019_CVPR}.
Several types of input representations for 3D local geometry have been explored, such as raw point cloud patches~\cite{Khoury_2017_ICCV,Deng_2018_CVPR}, voxel grids~\cite{Zeng_2017_CVPR,Gojcic_2019_CVPR} and multi-view images~\cite{Huang:2017:LLS:3151031.3137609,Zhou_2018_ECCV}.
Currently, on the geometric registration benchmark of 3DMatch~\cite{Zeng_2017_CVPR}, most learning-based methods are built upon either PointNet~\cite{Qi_2017_CVPR} with point cloud patches or 3D CNNs with voxel grids, and 3DSmoothNet~\cite{Gojcic_2019_CVPR} achieves the state-of-the-art performance with smoothed density value voxelization.
Despite the impressive progress made by the voxel representation, 
literature on 3D shape recognition and retrieval~\cite{Su_2015_ICCV,Qi_2016_CVPR,DBLP:conf/bmvc/WangPS17} indicates superior performance of multi-view images than voxel grids, and some initial attempts~\cite{Huang:2017:LLS:3151031.3137609,Zhou_2018_ECCV} have been made to extend a similar idea to 3D local descriptors.
Meanwhile, a line of recent studies has advanced 2D CNNs in learning local descriptors from a single image patch~\cite{Han:2015:MatchNet,Tian:2017:L2Net,Mishchuk:2017:HardNet,Yi:2016:LIFT,Keller_2018_CVPR,Mishkin_2018_ECCV,Luo_2018_ECCV}.
These motivate us to perform further investigation into a multi-view representation for 3D points and their local geometry.

The main challenges of adopting the multi-view representation in learning descriptors are as follows.
First, to obtain multi-view images, a set of viewpoints (virtual cameras) are needed for 3D graphics rendering pipelines in a preprocessing stage~\cite{Su_2015_ICCV,Huang:2017:LLS:3151031.3137609}.
In existing studies~\cite{Su_2015_ICCV,Qi_2016_CVPR,Huang:2017:LLS:3151031.3137609,DBLP:conf/bmvc/WangPS17,Feng_2018_CVPR,He:2019:NGram}, the viewpoints are either randomly sampled or heuristically hand-picked.
However, how to determine the viewpoints in a {data-driven manner} to produce more informative renderings for neural networks still remains a question.
Second, an effective fusion operation is required to integrate features from multiple views into a single compact descriptor.
Max-view pooling is a dominant fusion approach~\cite{Su_2015_ICCV,Qi_2016_CVPR,Huang:2017:LLS:3151031.3137609,DBLP:conf/bmvc/WangPS17},
but this operation might overlook subtle details~\cite{Zhou_2018_ECCV,DBLP:conf/bmvc/WangPS17}, leading to sub-optimal performance.

\begin{figure*}[ht]
	\centering
	\includegraphics[width=0.9\linewidth]{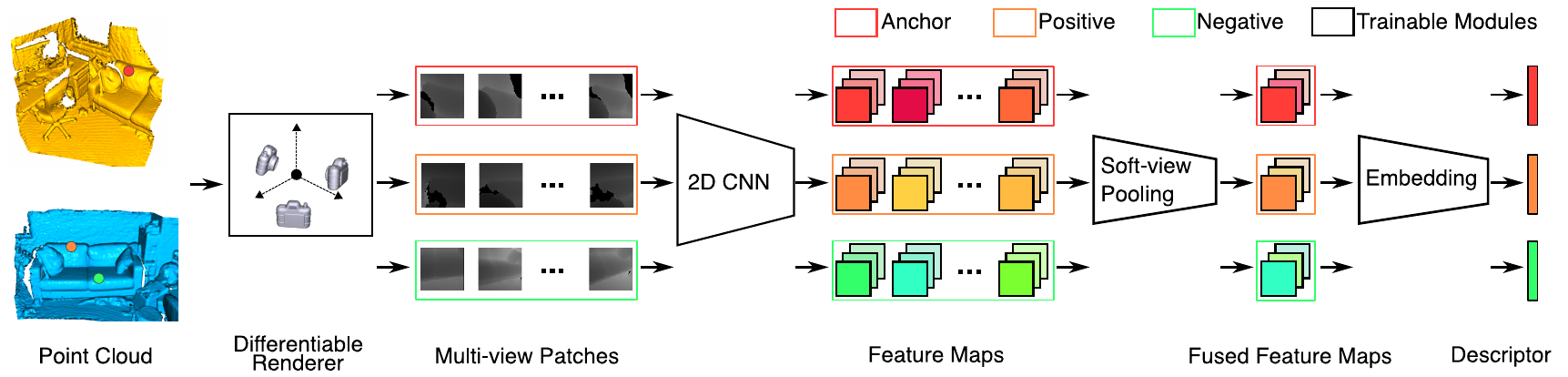}
	\caption{An end-to-end network that learns local multi-view descriptors for point clouds.
		The network takes point clouds as input and performs in-network multi-view rendering with a differentiable renderer for points of interest. Feature maps are extracted individually from each view and fused together via a soft-view pooling module to obtain the final descriptors.}
	\label{fig:pipeline}
\end{figure*}

In this work, we propose a novel network architecture that learns local multi-view descriptors for 3D point clouds in an end-to-end manner, as illustrated in \Figure{\ref{fig:pipeline}}.
Our network consists of three main stages: (1) multi-view rendering for a 3D point of interest of a point cloud; (2) feature extraction in each rendered view; and (3) feature fusion across the views.
Specifically, we first use an in-network differentiable renderer~\cite{Liu:2019:SR}
to project the 3D local geometry of a specific point as multi-view patches.
Viewpoints used by the renderer are optimizable parameters during training.
The renderer can back-propagate supervision signals from rendered pixels to the viewpoints, enabling joint optimization of the rendering stage with the other two stages.
Next, to extract features in each rendered view, we leverage existing CNNs that are well matured in the task of learning single patch descriptors~\cite{Tian:2017:L2Net,Mishchuk:2017:HardNet}.
Lastly, to fuse the features across all the views, we examine the gradient flow problem of max-view pooling~\cite{Su_2015_ICCV} and then design a novel soft-view pooling module.
The former only considers the strongest response across views for each position in feature maps, while in contrast, our design adaptively aggregates all the responses with attentive weights estimated by a sub-network.
In the backward pass, our design allows supervision signals to better flow into each input view for optimization.
The experiments conducted on the 3DMatch benchmark~\cite{Zeng_2017_CVPR} shows that our method outperforms existing hand-crafted and learned descriptors, and is robust against rotation and point density as well.

Our contributions in this work are summarized as:
(1) we propose 
a novel end-to-end framework for learning local multi-view descriptors of 3D point clouds, with the state-of-the-art performance;
(2) the viewpoints are optimizable via in-network differentiable rendering;
(3) a soft-view pooling module fuses features across views attentively with a better gradient flow.
We will make our code publicly available.

\section{Related Work}
\label{sec:related-work}

\textbf{Hand-crafted 3D Local Descriptors.}
Over the past few decades, a large body of literature has investigated descriptors for encoding geometric information of local neighborhoods of 3D points. 
A full review is beyond the scope of this paper.
Classic descriptors include, to name a few, 
Spin Image~\cite{Johnson:1999:SPIN}, 
3D Shape Contexts~\cite{Frome:2004:3DShapeContext}, 
PFH~\cite{Rusu:2008:PFH}, 
FPFH~\cite{Rusu:2009:FPFH},
SHOT~\cite{Tombari:2010:USH},
and Unique Shape Context~\cite{Tombari:2010:USC:1877808.1877821}.
These hand-crafted descriptors are mostly constructed from histograms of low-level geometric properties.
Despite the progress made by these descriptors, they may fail to handle well the nuisances commonly observed in real scanned data, like noise, incompleteness, and low resolution~\cite{Guo:2016:CPELFD}.

\textbf{Learned 3D Local Descriptors.}
With the recent success of deep neural networks~\cite{Russakovsky2015}, more attention has been shifted to developing learning-based 3D local descriptors~\cite{Deng_2018_CVPR,Khoury_2017_ICCV,Zeng_2017_CVPR,Gojcic_2019_CVPR,Huang:2017:LLS:3151031.3137609}.
In general, these methods fall into three categories according to input representations, including point cloud patches, voxel grids and multi-view images.

\emph{Point cloud patches} are the most straightforward representation for local neighborhoods of points.
PointNet, a seminal work done by \Etal{Qi}~\cite{Qi_2017_CVPR}, is specifically designed to handle the unstructured nature of point clouds.
Studies like~\cite{Deng_2018_CVPR,Deng_2018_ECCV,Xing:2018:3DTNet} build upon PointNet to learn descriptors for point cloud patches.
There also exist PointNet-based works that learn local descriptors jointly with other tasks, such as keypoint detection~\cite{yew2018-3dfeatnet} and pose prediction~\cite{Deng:2019:LFDPR}.

\emph{Voxel grids}, used in works like 3DMatch~\cite{Zeng_2017_CVPR} and 3DSmoothNet~\cite{Gojcic_2019_CVPR}, are a common structured representation for 3D point clouds~\cite{Maturana:2015:VoxNet,Wu:2015:3DSN,Qi_2016_CVPR}.
To reduce noise and boundary effects, \Etal{Gojcic}~\cite{Gojcic_2019_CVPR} proposed to use smoothed density value voxelization in 3DSmoothNet.
Their method achieves the state-of-the-art performance on the 3DMatch benchmark~\cite{Zeng_2017_CVPR}, substantially outperforming the aforementioned PointNet-based approaches~\cite{Deng_2018_CVPR,Deng_2018_ECCV,Deng:2019:LFDPR}.

\emph{Multi-view images} have demonstrated better performance than voxel grids in the task of 3D shape recognition and retrieval~\cite{Su_2015_ICCV,Qi_2016_CVPR,Roveri_2018_CVPR}, owing to their ability of delivering rich information of 3D geometry.
Motivated by the success in global shape analysis, researchers have extended the multi-view representation to 3D local descriptor learning~\cite{Huang:2017:LLS:3151031.3137609,Zhou_2018_ECCV}.
\Etal{Huang}~\cite{Huang:2017:LLS:3151031.3137609} re-purposed the CNN architecture from~\cite{Su_2015_ICCV,Alex:NIPS2012:AlexNet} to extract local descriptors of 3D shapes (e.g., airplanes or chairs) from multi-view images, which are rendered offline with clustered viewpoints.
There exist studies like~\cite{Elbaz:2017:PCRL,Roveri_2018_CVPR} that use 2D filtering for in-network image generation from point clouds.
{In contrast, our work considers the viewpoints as optimizable parameters and performs multi-view rendering with a differentiable renderer~\cite{Liu:2019:SR} in neural networks.}

To fuse view features into a single compact representation,
max-view pooling is widely used owing to its computational efficiency and view-order invariance~\cite{Su_2015_ICCV,Qi_2016_CVPR,DBLP:conf/bmvc/WangPS17,Huang:2017:LLS:3151031.3137609,Roveri_2018_CVPR,Zhou_2018_ECCV}, but it tends to overlook subtle details as discussed in~\cite{DBLP:conf/bmvc/WangPS17,Zhou_2018_ECCV,Mishchuk:2017:HardNet,Matthew:2013:SPR,Murray_2014_CVPR}.
\Etal{Zhou}~\cite{Zhou_2018_ECCV} proposed Fuseption, a residual-learning module for feature fusion, but their module is not view-order invariant and its number of parameters grows with the number of input views.
Alternative approaches, such as feature aggregation with NetVLAD~\cite{Arandjelovic_2016_CVPR} and RNN~\cite{Han:2019:SeqView}, have also been explored, but excessive computation or view ordering is required.
Differently, by analyzing the gradient flow of max-view pooling, we propose soft-view pooling that adaptively aggregates features with attentive weights in a view-order invariant manner.

\textbf{Differentiable Rendering.}
The conventional 3D graphics rendering pipeline involves rasterization and visibility test, which are non-differentiable discretization operations with respect to the projected point coordinates and view-dependent depths~\cite{Liu:2019:SR}.
Thus supervision signals cannot flow from the 2D image space to the 3D shape space,  preventing the integration of this pipeline into neural networks for end-to-end learning.
Recently researchers have designed several differentiable rendering frameworks~\cite{Loper:2014:OpenDR,kato2018renderer,Liu:2018:PAS:3272127.3275047,Li:2018:SketchR2CNN,Felix:2019:Pix2Vex,Wang:2019:DSS,Chen:2019:LP3DO,Liu:2019:SR} that incorporate approximated gradient formulations for the discretization operations.
Among them,
Soft Rasterizer (SoftRas), a state-of-the-art differentiable renderer developed by \Etal{Liu}~\cite{Liu:2019:SR}, treats mesh rendering as a process of probabilistic aggregation of triangles.
In this work, we modify SoftRas to extend its application to point cloud rendering and adopt a hard-forward soft-backward scheme.

\section{Methodology}
\label{sec:method}

Given a 3D point cloud $\PointCloudP$, we aim at training a neural network $\MVDescExtractor$ that can extract a discriminative local descriptor for a point $\PointP \in \PointCloudP$ in an end-to-end manner.
To this end, we perform projective analysis on the local geometry of $\PointP$ by using a multi-view representation.
Compared to point cloud patches or voxel grids, the multi-view representation can capture different levels of local context more easily~\cite{Huang:2017:LLS:3151031.3137609,Qi_2016_CVPR}.

Our network $\MVDescExtractor$ is comprised of three stages as shown in \Figure{\ref{fig:pipeline}}.
{First, the network $\MVDescExtractor$ directly takes the point cloud $\PointCloudP$ and the point of interest $\PointP$ as inputs and employs SoftRas~\cite{Liu:2019:SR} to render the local neighborhood of $\PointP$ as multi-view patches (\Section{\ref{subsec:multi-view-rendering}}).}
Second, we extract convolutional feature maps from each rendered view patch through a lightweight 2D CNN (\Section{\ref{subsec:feature-extraction}}).
Lastly, all the extracted view features are compactly fused together by a novel soft-view pooling module to obtain the local descriptor (\Section{\ref{subsec:multi-view-fusion}}).
The three stages of $\MVDescExtractor$ are jointly trained in an end-to-end manner such that descriptors of corresponding points that are geometrically and semantically similar are close to each other, while descriptors of non-corresponding points are distant to each other (\Section{\ref{subsec:training}}).

\subsection{Multi-view Rendering}
\label{subsec:multi-view-rendering}

\begin{figure}[ht]
	\centering
	\includegraphics[width=0.8\linewidth]{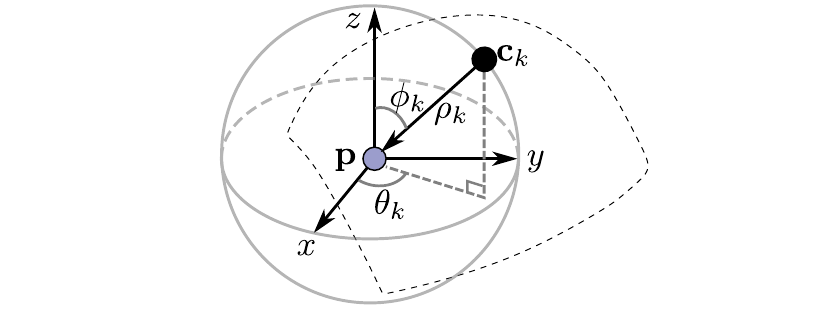}
	\caption{Local spherical coordinates ($\ViewPointTheta_k$, $\ViewPointPhi_k$, $\ViewPointRadius_k$) for a viewpoint $\ViewPoint_k$.}
	\label{fig:lrf}
\end{figure}

\textbf{Optimizable Viewpoints.}
Existing multi-view approaches select a set of rendering viewpoints according to certain rules, 
{e.g., by clustering~\cite{Huang:2017:LLS:3151031.3137609} or circling around a viewing center at a fixed step~\cite{Su_2015_ICCV,DBLP:conf/bmvc/WangPS17,Feng_2018_CVPR}.}
However, this view selection process is detached from the subsequent multi-view fusion stage, and thus might produce less representative inputs for the latter.
SoftRas allows the viewpoints to be optimizable parameters, which can be jointly trained with other network parameters in later stages.
To set up virtual cameras in a \emph{look-at} manner~\cite{Angel:2011:ICG:2018863}, 
we define the viewpoint parameters as $\{\ViewPoint_k = (\ViewPointTheta_k, \ViewPointPhi_k, \ViewPointRadius_k, \ViewPointUp)\}_{k=1}^{\ViewNum}$ using spherical coordinates, where $\ViewNum$ is the number of viewpoints.
Each viewpoint $\ViewPoint_k$ is represented by two angles $\ViewPointTheta_k$ and $\ViewPointPhi_k$, the distance $\ViewPointRadius_k$ from the local origin and a consistent upright orientation $\ViewPointUp$.
Given the point of interest $\PointP$ as the origin, the local reference frame (LRF) for $\{\ViewPoint_k\}$ is defined as follows (\Figure{\ref{fig:lrf}}):
the $z$-axis is collinear to the normal of $\PointP$;
the $x$-axis is the cross product of $\ViewPointUp$ and the $z$-axis {(a small perturbation to $\ViewPointUp$ if the normal is parallel to $\ViewPointUp$)};
and the $y$-axis is the cross product of the $z$-axis and $x$-axis.
We constrain $\{\ViewPoint_k\}$ to be within the hemisphere where the point normal resides (\Section{\ref{subsec:training}}).
To augment rotation invariance in the learned descriptors, we rotate each rendered view patch at 90-degree intervals~\cite{Huang:2017:LLS:3151031.3137609} (i.e., 4 in-plane rotations) within the network. 
Thus, a set of $4 \ViewNum$ view patches are obtained through rendering as detailed next.

\begin{figure}[ht]
	\centering
	\includegraphics[width=0.95\linewidth]{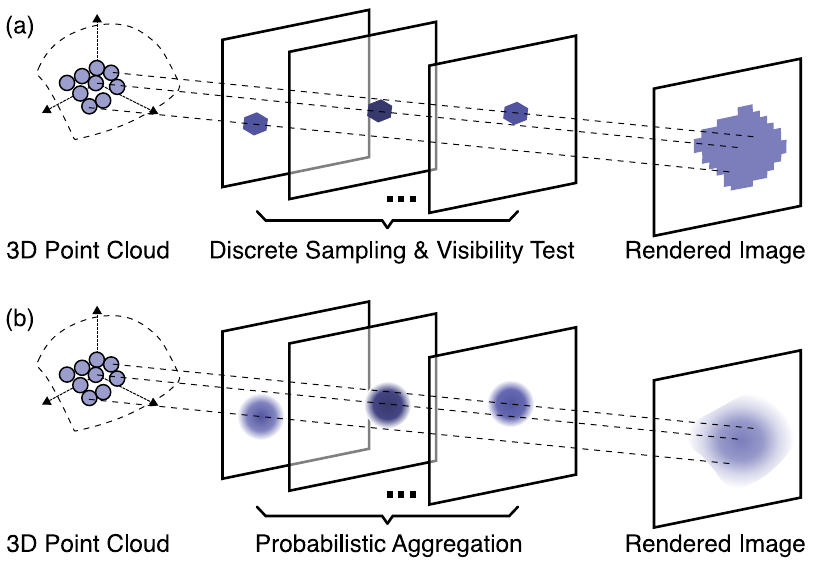}
	\caption{Rendering pipelines for point clouds: (a) Conventional 3D graphics rendering;
		(b) Soft Rasterizer~\cite{Liu:2019:SR} extended to 3D point cloud rendering.} 
	\label{fig:std_diff_rendering}
\end{figure}

\textbf{Differentiable Rendering.}
To address the non-differentiable issue of the conventional 3D graphics rendering pipeline (\Figure{\ref{fig:std_diff_rendering}-a}), SoftRas treats mesh rendering as a process of probabilistic aggregation of triangles in 2D.
To render the point cloud $\PointCloudP$ as view patches with $\{\ViewPoint_k\}$, one approach is to firstly transform $\PointCloudP$ to a mesh via surface reconstruction~\cite{Kazhdan:2006:PSR}, which, however, is challenging to integrate into our end-to-end framework and may not handle noise well (e.g., in laser scans of outdoor scenes).
Instead, we modify SoftRas to make it amenable to point cloud rendering (\Figure{\ref{fig:std_diff_rendering}-b}).
We consider each point $\PointQ_j \in \PointCloudP$ as a sphere~\cite{Huang:2017:LLS:3151031.3137609}, {whose radius can be a fixed value~\cite{Huang:2017:LLS:3151031.3137609}} or derived from the average distance between $\PointQ_j$ and its local neighbors.
After perspective projection with a specific viewpoint $\ViewPoint_k$, the point $\PointQ_j$ produces a probability map $\ProbabilityMap_j$ that describes the probability of each output pixel being covered by $\PointQ_j$~\cite{Liu:2019:SR}.
The $i$-th pixel in the rendering output $\ViewImage$ (of size 64 $\times$ 64) is defined as
\begin{equation}
\label{eq:softras}
\ViewImage^{i} = \sum_{j} \AggregationWeight(\ProbabilityMap_j^{i}, \PointDepth_j) \PointProperty_j + \AggregationWeight_b \PointProperty_b ,
\end{equation}
where $\PointProperty_j$ is the rendered attribute (e.g., color or view-dependent depth) of $\PointQ_j$, $\PointProperty_b$ is a default background value, and $\PointDepth_j$ is the depth of $\PointQ_j$.
The weighting function $\AggregationWeight(\cdot)$ designed in~\cite{Liu:2019:SR} is biased to points that are closer to the camera and the $i$-th pixel, and $\sum_{j} \AggregationWeight(\cdot) + \AggregationWeight_b = 1$.
Such a linear formulation in \Equation{\ref{eq:softras}} approximates the rasterization and visibility test in the conventional rendering pipeline (\Figure{\ref{fig:std_diff_rendering}}), and it is naturally differentiable.
Since input point clouds may lack color information, we use view-dependent depth as $\PointProperty_j$~\cite{Elbaz:2017:PCRL,Xie:2015:PFL}, which is invariant to illumination changes.
We refer the interested reader to~\cite{Liu:2019:SR} for detailed implementations and discussions of  $\ProbabilityMap_j$ and $\AggregationWeight(\cdot)$.

\begin{figure}[ht]
	\centering
	\includegraphics[width=\linewidth]{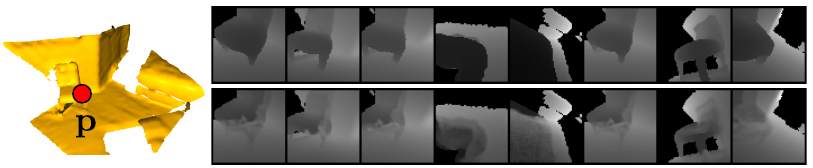}
	\caption{Multi-view rendering samples (depth, size = 64$\times$64) for a point $\PointP$. Top: renderings of our hard-forward soft-backward scheme (\Figure{\ref{fig:std_diff_rendering}}-a); Bottom: renderings of Soft Rasterizer~\cite{Liu:2019:SR} (\Figure{\ref{fig:std_diff_rendering}}-b).} 
	\label{fig:hard_soft_fwd}
\end{figure}

Although the differentiability of \Equation{\ref{eq:softras}} makes it possible for in-network rendering, we observed artifacts, such as blurry pixels at regions with large depth discontinuity, in the rendering outputs (see \Figure{\ref{fig:hard_soft_fwd}}).
To mitigate the influence of artifacts on the subsequent feature extraction, we instead adopt a hard-forward soft-backward scheme for rendering point clouds with SoftRas, sharing a similar idea to~\cite{kato2018renderer}.
Specifically, in the forward pass, we perform rasterization and visibility test to obtain rendering results in the same way as the conventional rendering pipeline (\Figure{\ref{fig:std_diff_rendering}-a}).
In the backward pass, we compute approximated gradients for the rendering using \Equation{\ref{eq:softras}} of SoftRas. 
We found that this approximation scheme works well in our experiments.

\subsection{Feature Extraction}
\label{subsec:feature-extraction}

Let $\{\ViewImage_k\}_{k=1}^{4\ViewNum}$ be the set of multi-view patches produced in the rendering stage for the point $\PointP$.
This 2D representation can naturally lend itself to existing patch analysis networks.
We adopt a lightweight CNN backbone similar to L2-Net~\cite{Tian:2017:L2Net,Mishchuk:2017:HardNet}, a state-of-the-art network for learning local image descriptors.
Concretely, the network is composed of six stacked convolutional layers, each followed by  normalization~\cite{UlyanovVL16:InstNorm} and ReLU layers.
We feed each patch $\ViewImage_k$ to the network and obtain a corresponding feature map denoted as $\ViewFeatureMap_k$, which is of size 8 $\times$ 8 with 128 channels.

\subsection{Multi-view Fusion}
\label{subsec:multi-view-fusion}

\begin{figure}[ht]
	\centering
	\includegraphics[width=0.8\linewidth]{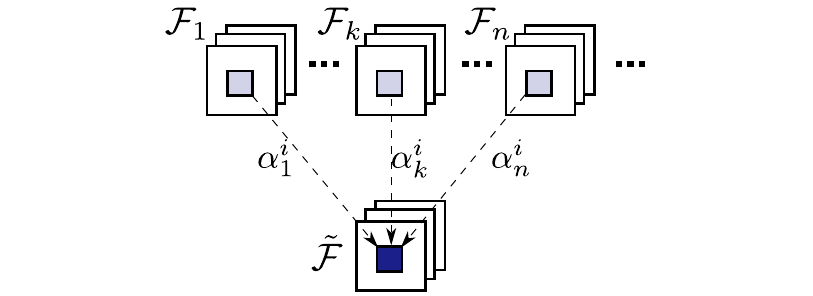}
	\caption{Multi-view fusion at location $i$ that iterates over all spatial and channel-wise positions (top: feature maps of each view; bottom: fused feature maps).}
	\label{fig:mvfusion}
\end{figure}

Given the set of feature maps $\{\ViewFeatureMap_k\}_{k=1}^{4\ViewNum}$ as input, we perform feature fusion across views to obtain a more compact multi-view representation.
Let $\FeatureFusion^{i}$ denote the feature value at location $i$ of the fused output $\FeatureFusion$ (the same size as $\ViewFeatureMap_k$), and $i$ iterates over all spatial and channel-wise positions (\Figure{\ref{fig:mvfusion}}).
Max-view pooling is a widely adopted fusion approach for its simple computation and invariance to view ordering.
However, this operation suffers from the following gradient flow problem in back-propagation.
Mathematically, max-view pooling can be expressed as
\begin{equation}
\label{eq:max-pool}
\FeatureFusion^{i} = \sum_{k} \FeatureFusionWeight_k^{i} \ViewFeatureMap_k^{i},
\end{equation}
where $\sum_{k} \FeatureFusionWeight_k^{i} = 1$ and the weights $\{\FeatureFusionWeight_k^{i}\}$ are in a one-hot form for selecting the maximum value.
In the backward pass, the gradient of \Equation{\ref{eq:max-pool}} is
\begin{equation}
\label{eq:max-pool-grad}
\dfrac{\partial \FeatureFusion^{i}}{\partial \ViewFeatureMap_k^{i}} = 
\begin{cases}
1 & \text{if } \ViewFeatureMap_k^{i} \text{ is the maximum value},\\
0 & \text{otherwise}.
\end{cases}
\end{equation}
Thus, according to the chain rule, supervision signals from loss functions cannot flow into certain locations in $\ViewFeatureMap_k$ if the locations do not have the maximum feature values, which may guide CNNs to overlook some details in feature extraction.
An alternative approach is average-view pooling with $\FeatureFusionWeight_k^{i} = \frac{1}{4\ViewNum}$ to alleviate the gradient flow problem.
However, as shown in existing studies~\cite{Huang:2017:LLS:3151031.3137609}, this approach performs worse than max-view pooling, partially because treating features equally across views may reduce the contribution of useful features while increasing the effect of insignificant features, leading to less discriminative descriptors.

Based on the above analysis, we propose soft-view pooling that adaptively estimates attentive weights $\{\FeatureFusionWeight_k^{i}\}$ with a sub-network.
Specifically, the sub-network takes each $\ViewFeatureMap_k$ as input and follows an encoder-decoder design to regress the corresponding weights.
The sub-network performs downsampling and then upsampling by a factor of 2 for both spatial size and channel depth, using a 3 $\times$ 3 convolutional layer and a 3 $\times$ 3 up-convolutional 
layer respectively, and a ReLU layer in-between.
The output weight map is denoted as $\FeatureFusionWeight_k$ (the same size as $\ViewFeatureMap_k$).
Afterward, for each location $i$ as defined above, the softmax function is applied to $\{\FeatureFusionWeight_k^{i}\}$ for normalization so that $\sum_{k} \FeatureFusionWeight_k^{i} = 1$ holds.
Note that the above computation is invariant to view orders.

At last, the network $\MVDescExtractor$ embeds the fused feature $\FeatureFusion$ to a $\MVDescDim$-dimensional descriptor space with a fully-connected layer and a subsequent $l2$ normalization layer.

\subsection{Training}
\label{subsec:training}

To train the network $\MVDescExtractor$, we sample matching point pairs in the overlapped region of two point clouds (at least 30\% overlap).
Given a batch of matching point pairs $\PointPairSet = \{(\PointP_i, \PointQ_i)\}$, we follow~\cite{Gojcic_2019_CVPR,Hermans:2017:BH} to adopt a batch-hard (BH) triplet loss
\begin{equation}
\label{eq:loss-batch-hard}
\begin{split}
\Loss_{BH} = \frac{1}{|\PointPairSet|}\sum_{i=1}^{|\PointPairSet|} \big[ \TripletMargin + \| \MVDescExtractor(\PointP_i) - \MVDescExtractor(\PointQ_i) \|_{2} - \\
\min_{\substack{j=1 \cdots |\PointPairSet| \\ j \ne i}} \| \MVDescExtractor(\PointP_i) - \MVDescExtractor(\PointQ_j) \|_{2} \big]_{+},
\end{split}
\end{equation}
where $[\cdot]_{+} = \max(\cdot, 0)$, and $\TripletMargin$ is a margin and set to 1.
For a training triplet, $\PointQ_i$ is the positive sample of $\PointP_i$, and $\Loss_{BH}$ considers the hardest negative sample $\PointQ_j$
within the batch $\PointPairSet$ for $\PointP_i$.
{As mentioned in \Section{\ref{subsec:multi-view-rendering}},}
we also impose range constraints for the optimizable viewpoints as follows:
\begin{equation}
\label{eq:loss-viewpoints}
\begin{split}
\Loss_{OV} &= \frac{1}{\ViewNum} \sum_{k=1}^{\ViewNum} \smashoperator[r]{\sum_{x \in \{ \ViewPointTheta_k, \ViewPointPhi_k, \ViewPointRadius_k \}}} [ |x - \frac{x_{a} + x_{b}}{2}| - \frac{x_{b} - x_{a}}{2} ]_{+},
\end{split}
\end{equation}
where $x_{a} = \{0,\ 0,\ 0.3\}$ and $x_{b} = \{2\pi,\ \pi/2,\ 1\}$ for $\ViewPointTheta_k$, $\ViewPointPhi_k$ and $\ViewPointRadius_k$ respectively.
Thus, the total loss is $\Loss = \Loss_{BH} + \lambda\Loss_{OV}$, where $\lambda$ is empirically set to 1.

We implemented the network with PyTorch~\cite{Pytorch:NIPS2019}.
We set the viewpoint number $\ViewNum = 8$ and the descriptor dimension $\MVDescDim = 32$ (\Section{\ref{subsec:ablation}}).
{The viewpoint parameters $\ViewPointTheta_k$, $\ViewPointPhi_k$, and $\ViewPointRadius_k$ were initialized randomly within the range in \Equation{\ref{eq:loss-viewpoints}}, and $\ViewPointUp$ was initialized to $[0, -1, 0]^{\top}$.}
We use Adam~\cite{KingmaB14:Adam} for stochastic gradient descent with $|\PointPairSet|=24$ and an initial learning rate of 0.001.
The network is trained for 16 epochs, and the learning rate is decayed by 0.1 every 4 epochs.

\section{Experiments}
\label{sec:experiments}

\subsection{3DMatch Benchmark}
\label{subsec:dataset}

\begin{table*}[ht]
	\centering
	\resizebox{0.95\textwidth}{!}{%
		\begin{tabular}{l|cc|cc|cc|cc|cc|cc|cc|cc|cc}
			\toprule
			& \multicolumn{2}{c|}{FPFH}& \multicolumn{2}{c|}{SHOT}& \multicolumn{2}{c|}{3DMatch}& \multicolumn{2}{c|}{CGF} & \multicolumn{2}{c|}{PPFNet}& \multicolumn{2}{c|}{PPF-FoldNet}& \multicolumn{2}{c|}{3DSmoothNet}  & \multicolumn{2}{c|}{LMVCNN} & \multicolumn{2}{c}{Ours}       \\
			$\tau_2$& 0.05        & 0.2        & 0.05      & 0.2          & 0.05        & 0.2           & 0.05        & 0.2        & 0.05        & 0.2          & 0.05        & 0.2               & 0.05          & 0.2               & 0.05          & 0.2         & 0.05            & 0.2          \\
			\midrule                                                                                                                                                                                                                                           
			Kitchen & 50.2        & 8.7        & 74.3      & 26.1         & 58.1        & 9.7           & 61.3        & 12.3       & 89.7        & -            & 78.7        & -                 & 97.4          & 62.8              & 98.8          & 76.5        & \textbf{99.4}  & \textbf{89.5} \\
			Home 1  & 70.5        & 23.1       & 80.1      & 48.7         & 72.4        & 17.3          & 72.4        & 23.7       & 55.8        & -            & 76.3        & -                 & 96.2          & 76.9              & 97.4          & 78.8        & \textbf{98.7}  & \textbf{85.9} \\
			Home 2  & 60.1        & 24.0       & 70.7      & 37.5         & 61.5        & 17.8          & 58.2        & 23.1       & 59.1        & -            & 61.5        & -                 & 90.9          & 66.3              & 90.9          & 68.3        & \textbf{94.7}  & \textbf{81.3} \\
			Hotel 1 & 71.2        & 6.2        & 77.4      & 26.5         & 54.4        & 0.9           & 62.8        & 8.8        & 58.0        & -            & 68.1        & -                 & 96.5          & 78.8              & \textbf{99.6} & 91.6        & \textbf{99.6}  & \textbf{95.1} \\
			Hotel 2 & 57.7        & 5.8        & 72.1      & 18.3         & 48.1        & 6.7           & 56.7        & 5.8        & 57.7        & -            & 71.2        & -                 & 93.3          & 72.1              & 99.0          & 90.4        & \textbf{100.0} & \textbf{92.3} \\
			Hotel 3 & 75.9        & 11.1       & 85.2      & 31.5         & 61.1        & 1.9           & 83.3        & 18.5       & 61.1        & -            & 94.4        & -                 & 98.1          & 88.9              & \textbf{100.0}& 90.7        & \textbf{100.0} & \textbf{94.4} \\
			Study   & 46.9        & 0.3        & 64.0      & 6.2          & 51.7        & 2.4           & 44.9        & 2.4        & 53.4        & -            & 62.0        & -                 & 94.5          & 72.3              & 95.2          & 77.4        & \textbf{95.5}  & \textbf{80.1} \\
			MIT Lab & 44.2        & 1.3        & 62.3      & 20.8         & 50.6        & 5.2           & 45.5        & 3.9        & 63.6        & -            & 62.3        & -                 & \textbf{93.5} & 64.9              & 90.9          & 74.0        & 92.2           & \textbf{76.6} \\
			\midrule                                                                                                                                                                                                                                           
			Average & 59.6        & 10.1       & 73.3      & 26.9         & 57.3        & 7.7           & 60.6        & 12.3       & 62.3        & -            & 71.8        & -                 & 95.0          & 72.9              & 96.5          & 81.0        & \textbf{97.5}  & \textbf{86.9} \\
			\bottomrule
		\end{tabular}
	}
	\caption{Average recall (\%) {of different methods on the 3DMatch benchmark with $\tau_1 = 10$cm and $\tau_2 = 0.05$ or $0.2$}.}
	\label{tab:3dmatch_recall}
\end{table*}

\textbf{Dataset.}
We evaluate the proposed method on the widely adopted geometric registration benchmark from 3DMatch~\cite{Zeng_2017_CVPR}.
The benchmark consists of RGB-D scans of 62 indoor scenes, an ensemble of several existing RGB-D datasets~\cite{ValentinDNKTIK16,Shotton:2013:SCRF,Xiao_2013_ICCV,Lai:2014:UFL,Halber2016StructuredGR}.
The data is split into 54 scenes for training and validation, and 8 scenes for testing.
In each scene, point cloud fragments are obtained by fusing 50 consecutive depth frames.
For each fragment in the testing set, a set of 5,000 randomly sampled points is provided as keypoints for descriptor extraction.

\textbf{Metric.}
The recall metric is used for comparisons on the testing set by averaging the number of matched point cloud fragments~\cite{Deng_2018_CVPR,Deng_2018_ECCV,Gojcic_2019_CVPR}.
Consider a set of point cloud fragment pairs $\FragmentPairSet = \{ ( \PointCloudP, \PointCloudQ )\}$, where point clouds $ \PointCloudP$ and $\PointCloudQ$ have at least 30\% overlap after alignment.
For a specific descriptor extraction method $\DescExtractor (\cdot)$, the set of putative matching points between $\PointCloudP$ and $\PointCloudQ$ is computed in the descriptor space as follows:
\begin{equation}
\label{eq:3dmatch-metric-point-corr}
\begin{split}
\MatchPointPairSet = \{ (\PointP \in \PointCloudP, \PointQ \in \PointCloudQ) | \DescExtractor(\PointP) = \text{nn}(\DescExtractor(\PointQ), \DescExtractor(\PointCloudP)) \wedge \\
\DescExtractor(\PointQ) = \text{nn}(\DescExtractor(\PointP), \DescExtractor(\PointCloudQ)) \},
\end{split}
\end{equation}
where $\PointP$ and $\PointQ$ are keypoints and $\text{nn} (\cdot)$ is the nearest neighbor search.
The recall metric $\MatchRecall$ is then defined as follows:
\begin{equation}
\label{eq:3dmatch-metric-recall}
\begin{split}
\MatchRecall = \dfrac{1}{|\FragmentPairSet|} \sum_{i=1}^{|\FragmentPairSet|}  \Big[\big(\dfrac{1}{|\MatchPointPairSet_i|} \sum_{\PointP, \PointQ \in \MatchPointPairSet_i} \big[\| \PointP - \MatchTransform_i (\PointQ) \|_2 < \tau_1 \big] \big) > \tau_2 \Big],
\end{split}
\end{equation}
where $[\cdot]$ is the Iverson bracket, and  $\MatchTransform_i (\cdot)$ is the ground-truth transformation for aligning the $i$-th fragment pair in $\FragmentPairSet$.
The distance threshold $\tau_1$ for matching points is set to 10 cm.
The inlier ratio $\tau_2$ ranges from 0.05 to 0.2.
To reliably find correct alignment parameters between two overlapping point clouds, the number of RANSAC~\cite{Fischler:1981:RANSAC} iterations is 55,000 for $\tau_2=0.05$ and 860 for $\tau_2=0.2$~\cite{Deng_2018_CVPR,Gojcic_2019_CVPR}.

\subsection{Evaluation Results}
\label{subsec:quantitative-results}

Following~\cite{Deng_2018_CVPR,Deng_2018_ECCV,Gojcic_2019_CVPR}, we compare our method (32-d) with several existing 3D local descriptors on the benchmark.
For hand-crafted descriptors, FPFH~\cite{Rusu:2009:FPFH} (33-d) and SHOT~\cite{Tombari:2010:USH} (352-d) are tested, and their implementations come from PCL~\cite{Rusu:2011:PCL}.
For learned descriptors, 3DMatch~\cite{Zeng_2017_CVPR} (512-d), CGF~\cite{Khoury_2017_ICCV} (32-d), PPFNet~\cite{Deng_2018_CVPR} (64-d), PPF-FoldNet~\cite{Deng_2018_ECCV} (512-d) and the current state-of-the-art 3DSmoothNet~\cite{Gojcic_2019_CVPR} (32-d) are tested.
{Additionally, we also compare with LMVCNN~\cite{Huang:2017:LLS:3151031.3137609}, a learned multi-view descriptor baseline using viewpoint clustering for offline rendering and max-view pooling for multi-view fusion.
	The original LMVCNN uses AlexNet~\cite{Alex:NIPS2012:AlexNet} as its CNN backbone and outputs 128-d descriptors, 
	but for fair comparisons, we reimplemented LMVCNN with the same CNN backbone
	and descriptor dimensionality (32-d) as our method.}
We use the implementations and trained weights from the authors for 3DMatch, CGF and 3DSmoothNet.
Since the implementations of PPFNet and PPF-FoldNet are not publicly accessible, we include their reported performance for completeness.

\Table{\ref{tab:3dmatch_recall}} shows the comparison results on the benchmark.
{For $\tau_2=0.05$, our method achieves an average recall of 97.5\%, outperforming all the competing descriptors.
	Nevertheless, $\tau_2=0.05$ is a relatively loose threshold on 3DMatch, since
	3DSmoothNet (95.0\%), LMVCNN (96.5\%) and our method all have achieved almost saturated performance with relatively small difference.
	Even so, our method obtains higher recalls in most testing scenes than 3DSmoothNet and LMVCNN.
	More notably, for a stricter condition $\tau_2=0.2$, there is significant improvement of our method over the other competitors.
	Specifically, our method maintains a high average recall of 86.9\%, while 3DSmoothNet and LMVCNN drop to 72.9\% and 81.0\%, respectively. 
	The performance of FPFH, SHOT, 3DMatch, and CGF falls below 30\%.}

\begin{figure}[ht]
	\centering
	\includegraphics[width=0.95\linewidth]{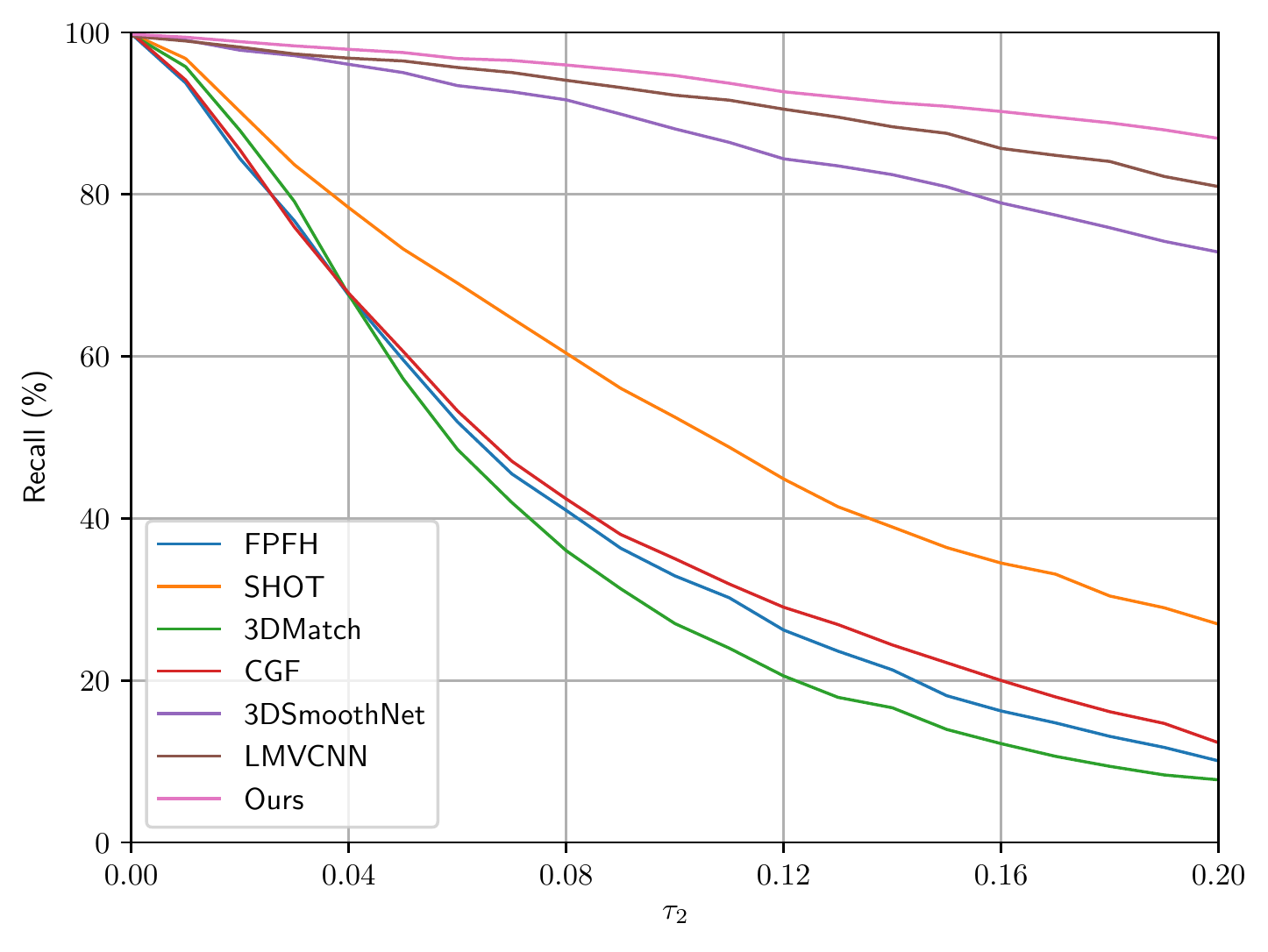}
	\caption{Average recall (\%) w.r.t inlier ratio $\tau_2$ on the 3DMatch benchmark.} 
	\label{fig:3dmatch_recall_varying_tau2}
\end{figure}

\begin{figure*}[ht]
	\centering
	\includegraphics[width=0.95\linewidth]{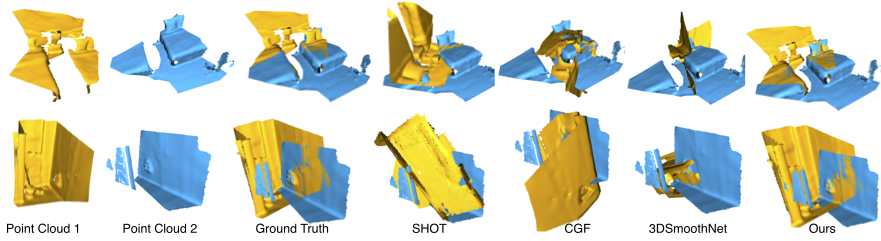}
	\caption{
		Geometric registration of point cloud 1 and point cloud 2 by different descriptors with RANSAC.} 
	\label{fig:3dmatch_geometric_registration}
\end{figure*}

In \Figure{\ref{fig:3dmatch_recall_varying_tau2}}, we plot the average recalls with respect to a range of $\tau_2$, 
illustrating the consistency of improvement brought by our method over the compared descriptors under different inlier ratio conditions.
Additionally, \Table{\ref{tab:3dmatch_corr_num}} lists the average number of correct correspondences found by each descriptor, which is computed as
$\frac{1}{|\FragmentPairSet|} \sum_{i=1}^{|\FragmentPairSet|} \sum_{\PointP, \PointQ \in \MatchPointPairSet_i} \big[\| \PointP - \MatchTransform_i (\PointQ) \|_2 < \tau_1 \big]$,
using the same notations as in \Equation{\ref{eq:3dmatch-metric-recall}}.
{It is observed that our multi-view descriptor is about 1.5$\times$ and 1.3$\times$ the average number of correspondences of 3DSmoothNet and LMVCNN, respectively.}
This clearly accounts for the dominant robustness of our descriptor.
{Additionally, \Figure{\ref{fig:3dmatch_geometric_registration}} visualizes some point cloud registration results obtained by different descriptors with RANSAC.
	Particularly, it is observed that our descriptor is robust
	in the registration of fragments with large flat regions (the second row).}

\begin{table}[ht]
	\centering
	\resizebox{\columnwidth}{!}{%
		\begin{tabular}{l|ccccccccc}
			\toprule
			& FPFH & SHOT & 3DMatch & CGF  & 3DSmoothNet & LMVCNN      & Ours          \\
			\midrule
			Kitchen & 104  & 154  & 104     & 131  & 274         & 276         & \textbf{380}  \\
			Home 1  & 158  & 207  & 134     & 168  & 325         & 344         & \textbf{438}  \\
			Home 2  & 132  & 183  & 125     & 159  & 318         & 314         & \textbf{395}  \\
			Hotel 1 & 103  & 131  & 74      & 95   & 272         & 347         & \textbf{457}  \\
			Hotel 2 & 105  & 124  & 64      & 101  & 239         & 286         & \textbf{407}  \\
			Hotel 3 & 131  & 160  & 65      & 134  & 277         & 301         & \textbf{446}  \\
			Study   & 65   & 84   & 66      & 58   & 172         & 239         & \textbf{299}  \\
			MIT Lab & 84   & 122  & 84      & 84   & 247         & 301         & \textbf{366}  \\
			\midrule
			Average & 110  & 146  & 89      & 116  & 266         & 301         & \textbf{398}  \\
			\bottomrule
		\end{tabular}
	}
	\caption{Average number of correct correspondences on the 3DMatch benchmark.}
	\label{tab:3dmatch_corr_num}
\end{table}


\textbf{Rotated 3DMatch Benchmark.}
{To evaluate the robustness of the descriptors against rotations},
we construct a rotated 3DMatch benchmark~\cite{Deng_2018_ECCV,Gojcic_2019_CVPR} by rotating the testing fragments with randomly sampled axes and angles in $[0, 2\pi]$.
The keypoint indices of each fragment are kept unchanged.
\Table{\ref{tab:3dmatch_rotated_recall}} gives the average recalls for each descriptor in the \emph{Rotated} column.
Our method achieves average recalls of 96.9\% and 82.1\% for $\tau_2 =$ 0.05 and 0.2 respectively, both surpassing the performance of 3DSmoothNet (94.9\% and 72.7\%), {LMVCNN (95.7\% and 76.7\%)} as well as the other descriptors.
The evaluation results indicate that our method can handle rotation well.

\textbf{Sparse 3DMatch Benchmark.}
To evaluate the robustness of the descriptors against point density, we follow~\cite{Deng_2018_ECCV,Gojcic_2019_CVPR} to construct a sparse 3DMatch benchmark.
Concretely, for each testing fragment, the keypoints are firstly retained and then 50\% or 25\% of the remaining points are randomly selected. 
The evaluation results are shown in
\Table{\ref{tab:3dmatch_rotated_recall}} (the last two columns).
It is found that owing to the {sphere-based rendering, our method is able to handle different point densities, like LMVCNN and 3DSmoothNet, and maintains the superior performance.}

\begin{table}[ht]
	\centering
	\resizebox{0.8\columnwidth}{!}{%
		\begin{tabular}{l|cc|cc|cc}
			\toprule
			& \multicolumn{2}{c|}{Rotated}         & \multicolumn{2}{c|}{Sparse (0.5)}  & \multicolumn{2}{c}{Sparse (0.25)} \\
			$\tau_2$        & 0.05              & 0.2              & 0.05              & 0.2            & 0.05              & 0.2           \\
			\midrule
			FPFH            & 60.1              & 10.0             & 59.2              & 9.5            & 57.8              & 8.5           \\
			SHOT            & 73.3              & 26.9             & 72.3              & 25.5           & 70.7              & 23.1          \\
			3DMatch         & 11.6              & 1.4              & 73.1              & 15.8           & 73.3              & 15.9          \\
			CGF             & 60.7              & 12.5             & 52.6              & 7.8            & 41.7              & 3.8           \\
			PPFNet          & 0.3               & -                & -                 & -              & -                 & -             \\
			PPF-FoldNet     & 73.1              & -                & -                 & -              & -                 & -             \\
			3DSmoothNet     & 94.9              & 72.7             & 94.4              & 71.7           & 94.8              & 70.1          \\
			LMVCNN          & 95.7              & 76.7             & 96.2              & 81.3           & 95.9              & 81.5          \\
			Ours            & \textbf{96.9}     & \textbf{82.1}    & \textbf{97.2}     & \textbf{87.2}  & \textbf{97.3}     & \textbf{86.1} \\
			\bottomrule
		\end{tabular}
	}
	\caption{Average recall (\%) on a rotated or sparse 3DMatch benchmark with $\tau_1 = 10$cm and $\tau_2 = 0.05$ or $0.2$.}
	\label{tab:3dmatch_rotated_recall}
\end{table}

\begin{table}[ht]
	\centering
	\resizebox{0.7\columnwidth}{!}{%
		\begin{tabular}{l|ccc}
			\toprule
			& Input prep.       & Inference        & Total           \\
			\midrule
			3DMatch         & 0.1               & 2.0              & 2.1             \\
			CGF             & 10.6              & 0.1              & 10.7            \\
			3DSmoothNet     & 39.4              & 0.2              & 39.6            \\
			Ours            & 7.2               & 1.5              & 8.7             \\
			\bottomrule
		\end{tabular}
	}
	\caption{Average running time (ms) per point on the 3DMatch benchmark.}
	\label{tab:3dmatch_running_time}
\end{table}

\textbf{Running Time.}
\Table{\ref{tab:3dmatch_running_time}} summarizes the running time for the learned descriptors on the standard 3DMatch benchmark.
All the experiments were performed on a PC with an Intel Core i7 @ 3.6GHz, a 32GB RAM and an NVIDIA GTX 1080Ti GPU.
The input preparation in \Table{\ref{tab:3dmatch_running_time}} refers to voxelization with TDF~\cite{Zeng_2017_CVPR} for 3DMatch, spherical histogram computation~\cite{Khoury_2017_ICCV} for CGF, LRF computation and SDV voxelization~\cite{Gojcic_2019_CVPR} for 3DSmoothNet, and multi-view rendering (\Section{\ref{subsec:multi-view-rendering}}) for our method.
The inference in \Table{\ref{tab:3dmatch_running_time}} refers to descriptor extraction from the prepared inputs with neural networks.
The results show that the input preparation stage dominates the running time of our method.
{Additionally, for sphere-based rendering (\Section{\ref{subsec:multi-view-rendering}}), it takes 0.16ms to determine a point radius by neighborhood query with FLANN~\cite{MujaL09:FLANN} (used in our implementation), while alternatively the computation can be eschewed by using a fixed radius as in~\cite{Huang:2017:LLS:3151031.3137609}.}
Nevertheless, our method still demonstrates competitive running time performance.

\subsection{Generalization to Outdoor Scenes}
\label{subsec:generalize-outdoor}

{We further evaluate the generalization ability of the descriptors on an outdoor-scene benchmark constructed by \Etal{Gojcic}~\cite{Gojcic_2019_CVPR} with point clouds from the ETH dataset~\cite{Pomerleau:2012}.
	{This} benchmark consists of four scenes, including Gazebo-Summer, Gazebo-Winter, Wood-Summer and Wood-Autumn.
	The point clouds were obtained by a laser scanner and mostly about outdoor vegetation.
	Thus, the point clouds are in {a} large spatial range with {a} low resolution and contain complex and noisy local geometry.
	Identical to the 3DMatch benchmark, 5,000 keypoints are randomly sampled in each point cloud for descriptor extraction.
	The evaluation metric is the same as {that} in \Section{\ref{subsec:dataset}}.
	Following~\cite{Gojcic_2019_CVPR}, no fine{-}tuning is performed for the descriptors trained on the 3DMatch benchmark.
	To accommodate the low resolution and large spatial range of the point clouds, the voxel grids for 3DMatch and 3DSmoothNet are enlarged with longer edges ($3\times$ and $5\times$ respectively) than those in \Section{\ref{subsec:quantitative-results}}.
	The radius of spherical histogram in CGF is $3.3\times$ longer.
	For LMVCNN and our method, the distance $\ViewPointRadius_k$ in each viewpoint $\ViewPoint_k$ is multiplied by a factor of 3.
}

The average recall results are shown in \Table{\ref{tab:eth_recall}}.
Our method (79.9\%) achieves comparable performance to 3DSmoothNet (79.0\%).
Meanwhile, our method significantly outperforms {LMVCNN (39.7\%)} and SHOT (61.1\%), and the other descriptors (including CGF, 3DMatch and FPFH) fall below 25\%.
{To account for the deteriorated performance of LMVCNN, further experiments on its used view selection and multi-view fusion strategies are performed in \Section{\ref{subsec:ablation}}.}
The above results show that our method trained on the 3DMatch benchmark can generalize well to outdoor scenes.

\begin{table}[ht]
	\centering
	\resizebox{0.75\columnwidth}{!}{%
		\begin{tabular}{l|cc|cc|c}
			\toprule
			& \multicolumn{2}{c|}{Gazebo}          & \multicolumn{2}{c|}{Wood}         &                \\
			& Sum.              & Wint.            & Sum.              & Aut.          & Avg.           \\
			\midrule
			FPFH            & 40.2              & 15.2             & 24.0              & 14.8          & 23.6           \\
			SHOT            & 73.9              & 45.7             & 64.0              & 60.9          & 61.1           \\
			3DMatch         & 22.8              & 8.7              & 22.4              & 13.9          & 16.9           \\
			CGF             & 38.6              & 15.2             & 19.2              & 12.2          & 21.3           \\
			3DSmoothNet     & \textbf{91.3}     & \textbf{84.1}    & 72.8              & 67.8          & 79.0           \\
			LMVCNN          & 53.3              & 31.8             & 42.4              & 31.3          & 39.7           \\
			Ours            & 85.3              & 72.0             & \textbf{84.0}     & \textbf{78.3} & \textbf{79.9}  \\
			\bottomrule
		\end{tabular}
	}
	\caption{Average recall (\%) on the ETH benchmark with $\tau_1 = 10$cm and $\tau_2 = 0.05$.}
	\label{tab:eth_recall}
\end{table}

\subsection{Ablation Study}
\label{subsec:ablation}

\begin{figure}[ht]
	\centering
	\includegraphics[width=0.95\linewidth]{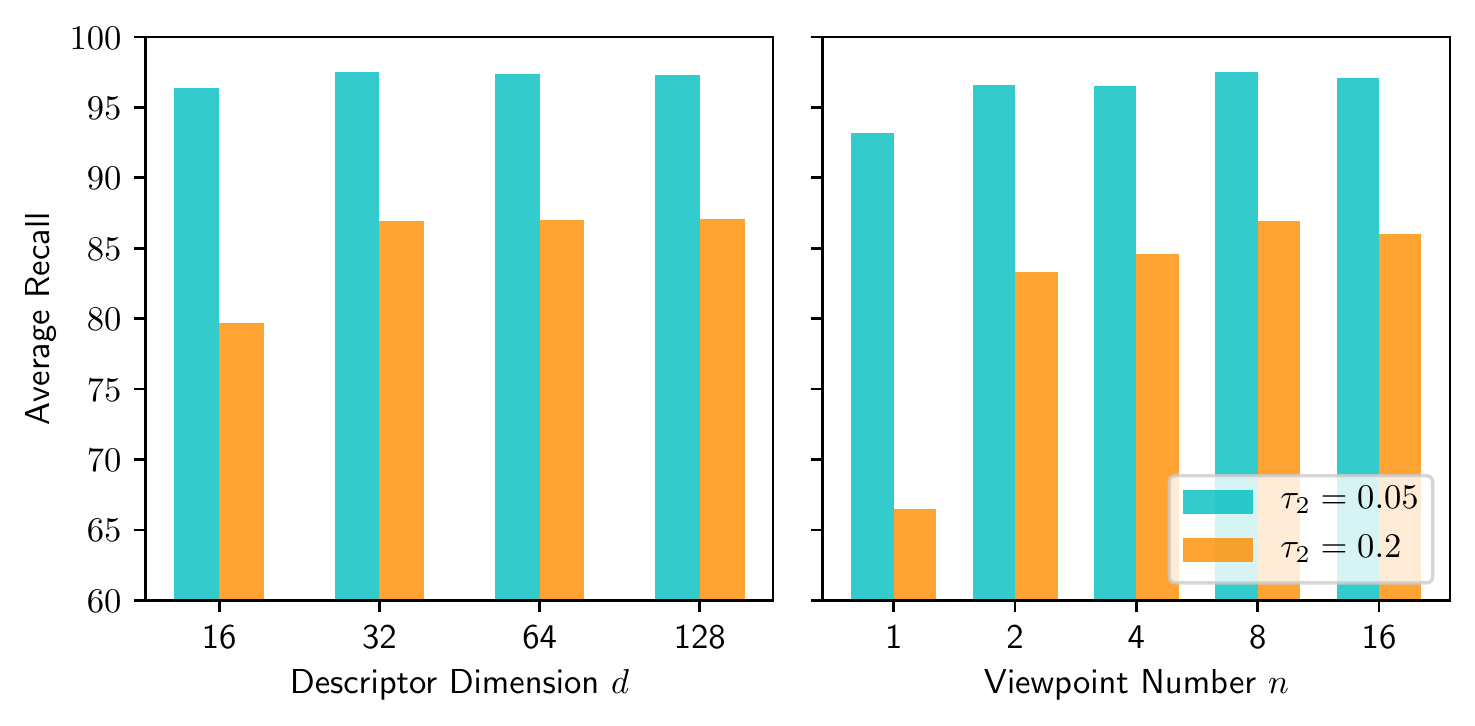}
	\caption{Average recall (\%) w.r.t descriptor dimension $\MVDescDim$ and viewpoint number $\ViewNum$ on the 3DMatch benchmark.} 
	\label{fig:3dmatch_featdim_viewnum}
\end{figure}

\textbf{Descriptor Dimension \& Viewpoint Number.}
In \Figure{\ref{fig:3dmatch_featdim_viewnum}} we plot the average recalls of our method with different descriptor dimensions $\MVDescDim$ and viewpoint numbers $\ViewNum$ (as defined in \Section{\ref{subsec:multi-view-fusion}} and \Section{\ref{subsec:multi-view-rendering}}).
It is found that increased descriptor dimensions ($\MVDescDim \geq 32$) and viewpoint numbers ($\ViewNum \geq 8$) lead to saturated performance.
Thus we adopt $\MVDescDim = 32$ and $\ViewNum = 8$ for our method in the experiments.

\textbf{Viewpoints.}
{In \Table{\ref{tab:3dmatch_ablation_02}} (top), we show the performance of our network $\MVDescExtractor$ trained with different viewpoint selection rules in multi-view rendering.
	Concretely, the straightforward \emph{random sampling} rule places the viewpoints randomly within the range in \Equation{\ref{eq:loss-viewpoints}}.
	The \emph{viewpoint clustering} rule used in LMVCNN~\cite{Huang:2017:LLS:3151031.3137609} selects three representative viewing directions via K-medoids clustering.
	The \emph{orbited placement} rule sets the viewpoints with $\ViewPointRadius=0.3$, $\ViewPointPhi=\pi/6$, and $\ViewPointTheta$ at a $\pi/4$ step (\Section{\ref{subsec:multi-view-rendering}}), similar to the strategy used in 3D shape recognition works~\cite{Su_2015_ICCV,DBLP:conf/bmvc/WangPS17,Feng_2018_CVPR}.
	The performance of $\MVDescExtractor$ without rotation augmentation to the rendered view patches is also provided.
	It is found that our optimizable viewpoints produce
	better performance than these alternative view selection rules, especially on the generalization ability to the ETH outdoor dataset.
}

\textbf{Multi-view Fusion.} 
{We perform experiments to compare our soft-view pooling with several alternative multi-view fusion approaches, including max-view pooling~\cite{Huang:2017:LLS:3151031.3137609}, Fuseption~\cite{Zhou_2018_ECCV}, and NetVLAD~\cite{Arandjelovic_2016_CVPR}. 
	We list the performance of the network $\MVDescExtractor$ trained with the above fusion approaches in \Table{\ref{tab:3dmatch_ablation_02}} (bottom).
	While on the 3DMatch dataset the improvement of soft-view pooling is small
	compared with max-view pooling,
	our method shows significantly better generalization on the ETH outdoor dataset.
	This is partially because the low-resolution scans of outdoor vegetation in ETH would produce relatively noisy renderings, presenting challenges to max-view pooling for selecting the strongest feature response.
	Differently, the response is adaptively gathered in our method with attention. 
}

\begin{table}[ht]
	\centering
	\resizebox{0.75\columnwidth}{!}{%
		\begin{tabular}{l|ccc}
			\toprule
			& \multicolumn{2}{c}{3DMatch}   & ETH             \\
			$\tau_2$                                                            & 0.05          & 0.2           & 0.05            \\
			\midrule
			Random sampling                                                     & 97.0          & 84.1          & 64.8            \\
			Viewpoint clustering                                                & 96.7          & 83.5          & 53.3            \\
			Orbited placement                                                   & 92.5          & 55.2          & 42.2            \\
			Ours w/o rotation augment.                                          & 96.9          & 85.6          & 54.9            \\
			Ours                                                                & \textbf{97.5} & \textbf{86.9} & \textbf{79.9}   \\
			\midrule
			Max-view pooling                                                    & 96.9          & 85.4          & 66.8            \\
			Fuseption                                                           & 97.1          & 85.1          & 55.9            \\
			NetVLAD                                                             & 95.9          & 77.4          & 58.7            \\
			Ours                                                                & \textbf{97.5} & \textbf{86.9} & \textbf{79.9}   \\
			\bottomrule
		\end{tabular}
	}
	\caption{Ablation study of viewpoint selection and multi-view fusion on the 3DMatch and ETH benchmarks.}
	\label{tab:3dmatch_ablation_02}
\end{table}

\section{Conclusion}
\label{sec:conclusion}

We have presented a novel end-to-end framework for learning local multi-view descriptors of 3D point clouds.
Our framework performs in-network multi-view rendering with optimizable viewpoints that can be jointly trained with later stages, and integrates convolutional features across views attentively via soft-view pooling.
We demonstrate the superior performance of our method and its generalization to outdoor scenes through experiments. 
For future work, it is worth investigating the acceleration of differentiable multi-view rendering of point clouds and the extension of our framework to other tasks such as 3D object detection and recognition in point clouds.

\section*{Acknowledgements}

This work was supported by grants from the Research Grants Council of the Hong Kong Special Administrative Region, China (Project No. CityU 11212119, {HKUST 16206819, HKUST 16213520}), and the Centre for Applied Computing and Interactive Media (ACIM) of School of Creative Media, CityU.

{\small
\bibliographystyle{ieee_fullname}
\bibliography{references}
}

\section{Supplementary Material}
\label{sec:supp-material}

\subsection{CNN}
\label{sec:supp-CNN}

In \Section{3.2} of the main text, we adopt a CNN architecture similar to L2-Net~\cite{Tian:2017:L2Net} to extract feature maps for each view patch.
The detailed configuration of the network is listed in \Table{\ref{tab:supp-network}}.
Note that the network input is of size 64$\times$64 with a single depth channel, and the final output is of size 8$\times$8 with 128 feature channels.

\begin{table}[ht]
	\centering
	\resizebox{\columnwidth}{!}{%
		\begin{tabular}{l|llcc}
			\toprule
			\#  & Layer                          & Kernel                & Stride    & Padding \\
			\midrule
			1   & Conv - Norm - ReLU           & 3$\times$3$\times$32  & 2         & 1       \\
			\midrule
			2   & Conv - Norm - ReLU           & 3$\times$3$\times$32  & 1         & 1       \\
			\midrule
			3   & Conv - Norm - ReLU           & 3$\times$3$\times$64  & 2         & 1       \\
			\midrule
			4   & Conv - Norm - ReLU           & 3$\times$3$\times$64  & 1         & 1       \\
			\midrule
			5   & Conv - Norm - ReLU           & 3$\times$3$\times$128 & 2         & 1       \\
			\midrule
			6   & Conv - Norm - ReLU           & 3$\times$3$\times$128 & 1         & 1       \\
			\bottomrule
		\end{tabular}
	}
	\caption{CNN backbone for feature extraction of each view patch.
		In the \emph{Kernel} column, the first two numbers represent the kernel size, and the third number is the number of output feature channels.}
	\label{tab:supp-network}
\end{table}

\subsection{Multi-view Rendering}
\label{sec:supp-multiview-rendering}

\begin{figure}[ht]
	\centering
	\includegraphics[width=\linewidth]{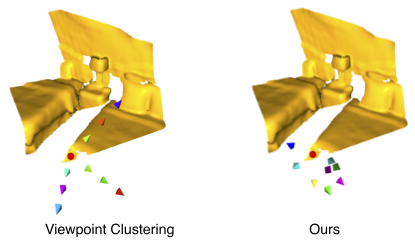}
	\caption{Visualization of viewpoints obtained by a clustering scheme and our method. The red spheres denote the points of interest, and the pyramids represent virtual cameras.}
	\label{fig:supp-viewpoints}
\end{figure}

In \Figure{\ref{fig:supp-viewpoints}}, we visualize the optimizable viewpoints after training.
We also show the viewpoints obtained by a clustering scheme similar to the one in~\cite{Huang:2017:LLS:3151031.3137609}.
Specifically, 150 spherical coordinates $(\ViewPointTheta, \ViewPointPhi)$ are randomly sampled on the hemisphere where  point normals reside, and then the k-medoids clustering algorithm is applied to select three viewing directions.
For each viewing direction, a virtual camera is placed at distances of 0.3m, 0.6m, 0.9m to the points of interest, and each rendered view patch is augmented with four in-plane rotations.

As shown in \Figure{\ref{fig:supp-viewpoints}}, there are mainly {two} differences between the hand-crafted rule and our method.
First, the hand-crafted rule places some viewpoints far from points of interest,
while the learnt viewpoints have {more concentrated distance range}, indicating the relatively low importance of broader global context. 
Second, the hand-crafted rule selects some dominant viewing directions {through clustering}, whereas the learnt viewpoints have {more distributed viewing directions} around the points of interests, which can help to capture more local geometry variance.
In sum, the learnt viewpoints effectively balance the extent of context-awareness and local details in extracted descriptors, challenging the design wisdom of hand-crafted rules.

\subsection{Multi-view Fusion}
\label{sec:supp-multiview-fusion}

In \Section{4.4} of the main text, we compared the proposed soft-view pooling with alternative fusion approaches including max-view pooling~\cite{Huang:2017:LLS:3151031.3137609,Su_2015_ICCV,Qi_2016_CVPR}, Fuseption~\cite{Zhou_2018_ECCV}, and NetVLAD~\cite{Arandjelovic_2016_CVPR}.
Fuseption has two branches: in the first branch, the feature maps of all the views are first channelwise concatenated together in a specific order and then fed into a convolutional block; in the second branch, max-pooling is applied to the inputs and the results are added to the output of the first branch, serving as a shortcut connection.
NetVLAD is a descriptor pooling method that summarizes the residuals of each input w.r.t. several learnable cluster centers.
The number of cluster centers is a hyper parameter, which is set to eight in our experiments.
The network $\MVDescExtractor$ is trained with the alternative fusion approaches, while the other stages are kept unchanged.
The descriptor dimension $\MVDescDim$ is set to 32, and the optimizable viewpoint number $\ViewNum$ is set to 8.

\begin{figure*}[ht]
	\centering
	\includegraphics[width=\linewidth]{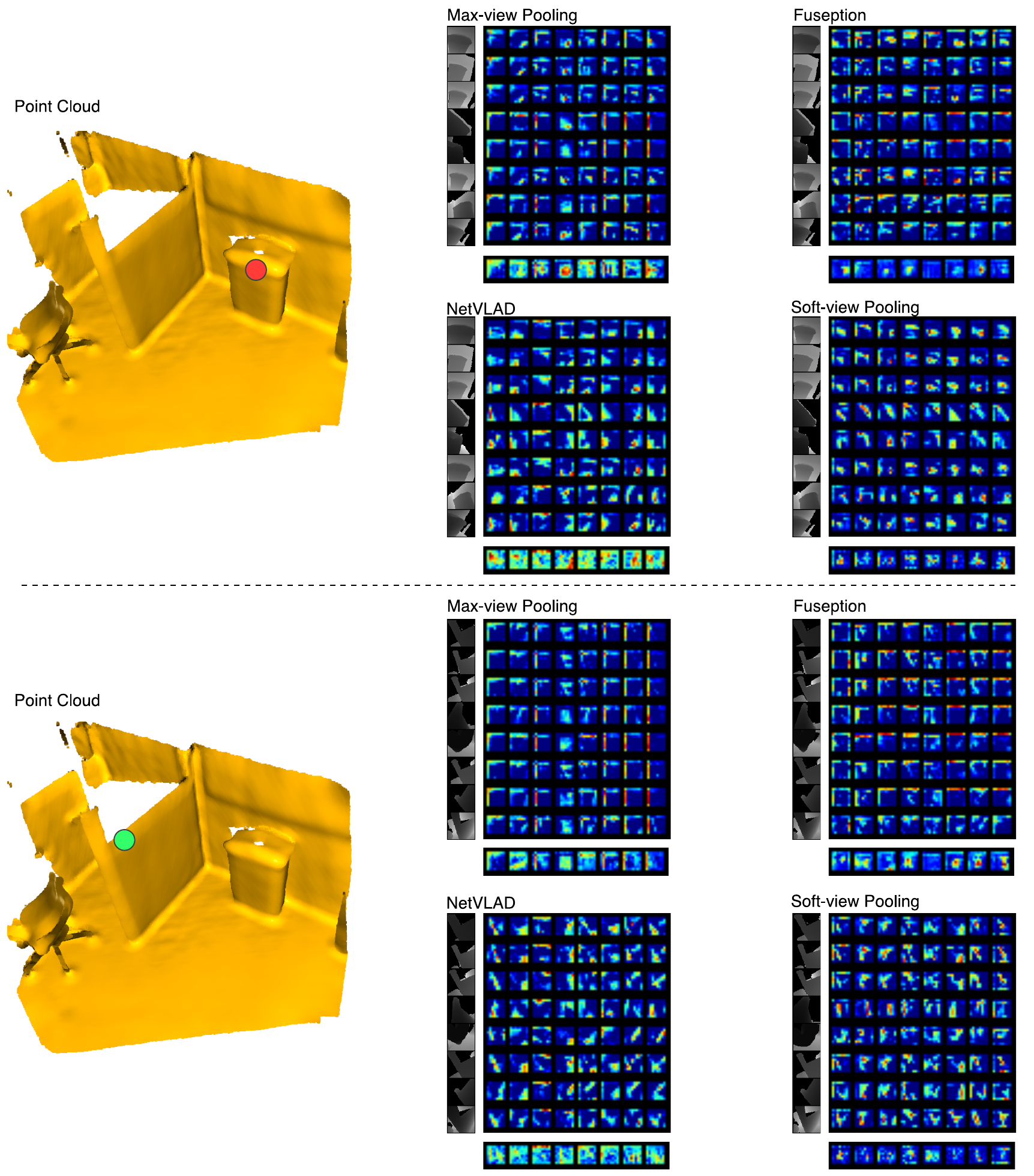}
	\caption{Visualizations for multi-view fusion by different methods. The top part is for the {\color{red}red} keypoint while the bottom part is for the {\color{green}green} keypoint. In each block, we visualize the view patches (depth) rendered with eight optimizable viewpoints on the left. On the right are the corresponding convolutional feature maps (with channel indices $\{1, 2, 4, 8, 16, 32, 64, 128\}$) before fusion, and each row is for a specific view. Fused feature maps across views are placed on the bottom.}
	\label{fig:supp-mvfusion_feature_maps}
\end{figure*}

In \Figure{\ref{fig:supp-mvfusion_feature_maps}}, we visualize the rendered multi-view inputs to CNNs, extracted feature maps for each view, and fused feature maps across views.
It is observed that the CNN is influenced by multi-view fusion for feature extraction.
Before fusion, for soft-view pooling and NetVLAD, the feature maps of each view extracted by the CNN tend to have more response, compared to max-view pooling and Fuseption.
After fusion, the feature maps produced by max-view pooling and NetVLAD tend to have more high response than soft-view pooling and Fuseption.
Note that for each location in the fused feature maps, max-view pooling only selects the strongest input response across views and discards the rest, while our soft-view pooling collectively considers all the inputs in an attentive manner for integration.

\subsection{Comparisons with 3DSmoothNet}
\label{sec:supp-compare-3dsmoothnet}

\begin{figure*}[ht]
	\centering
	\includegraphics[width=\linewidth]{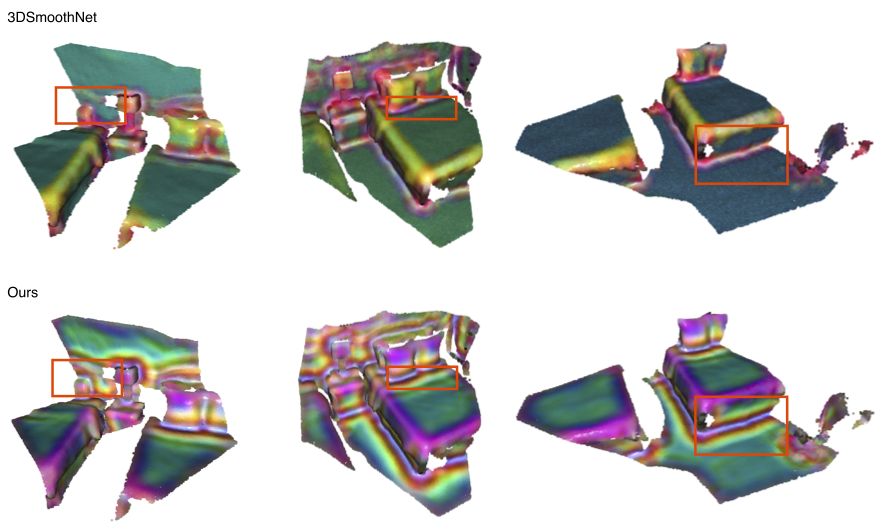}
	\caption{Visualization of local descriptors for 3DSmoothNet and our method. The high dimensional descriptors are projected with PCA to 3D space and color-coded. The highlighted regions show that our method can better capture geometric changes in the point clouds.}
	\label{fig:supp-desc_visualization}
\end{figure*}

\begin{figure*}[ht]
	\centering
	\includegraphics[width=\linewidth]{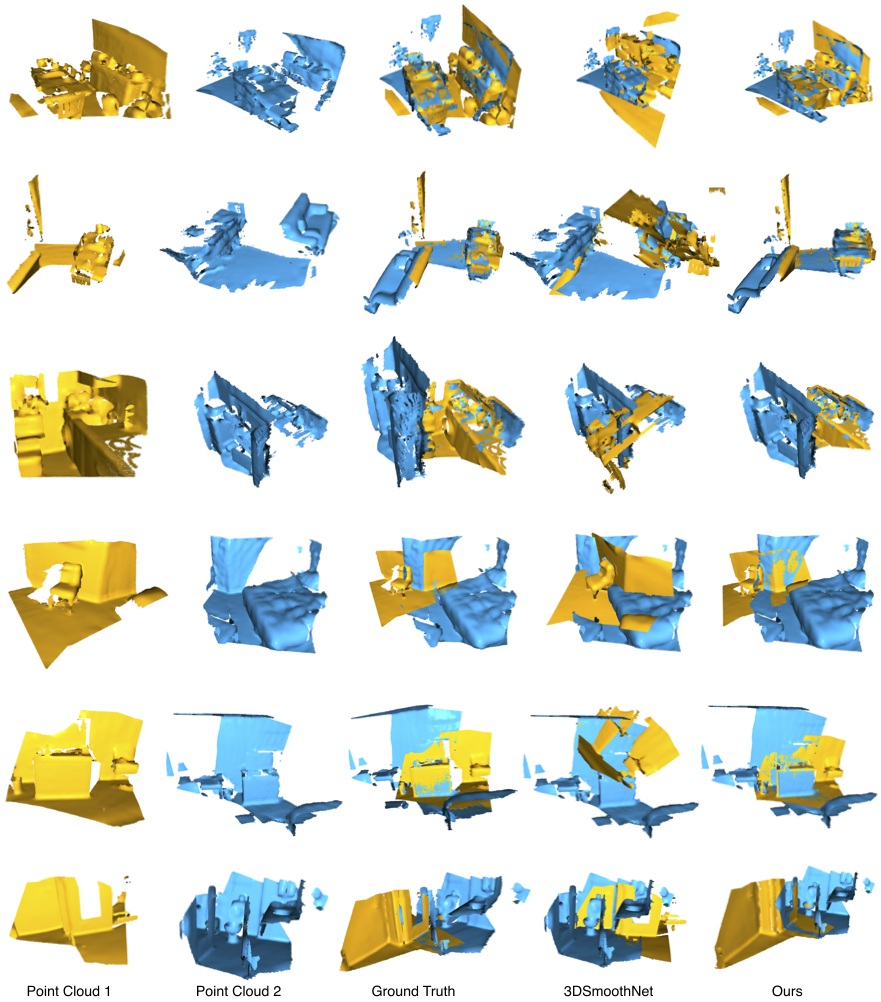}
	\caption{More geometric registration results with RANSAC for 3DSmoothNet and our method.}
	\label{fig:supp-3dmatch_registration}
\end{figure*}

In \Figure{\ref{fig:supp-desc_visualization}}, we visualize the color-coded local descriptors for all the points in the point clouds.
Specifically, we project the high dimensional descriptors with PCA and keep the first three components, which are color-coded.
It is observed that the descriptors of 3DSmoothNet and our method are both geometry-aware.
Particularly, our method is able to capture more geometric changes in the point clouds (see the highlighted wall, pillow and floor regions of the point clouds in \Figure{\ref{fig:supp-desc_visualization}}).
In \Figure{\ref{fig:supp-3dmatch_registration}}, we show additional geometric registration results of point cloud pairs, which further demonstrate the above advantage of our method.

For the running time of 3DSmoothNet in \Section{4.2} of the main text, we observed some gap between our experiment results (input prep: 39.4ms; inference: 0.2ms) and the performance reported by the authors (input prep: 4.2ms; inference: 0.3ms).
We used the source code\footnote{\url{https://github.com/zgojcic/3DSmoothNet}} of 3DSmoothNet released by the authors, and the running time gap of input preparation is likely due to the difference of hardware configurations.
In~\cite{Gojcic_2019_CVPR}, they used a PC with an Intel Xeon E5-1650, a 32GB RAM and an NVIDIA GeForce GTX 1080 GPU, while we used a PC with an Intel Core i7 @ 3.6GHz, a 32GB RAM and an NVIDIA GTX 1080Ti GPU.
Their input preparation stage involving LRF computation and SDV voxelization runs on CPU, which may be accelerated with GPU for further improvement.

\end{document}